\documentclass[final]{cvpr}

\usepackage{times}
\usepackage{epsfig}
\usepackage{graphicx}
\usepackage{amsmath}
\usepackage{amssymb}
\usepackage{appendix}

\usepackage{booktabs}

\usepackage{array,multirow,graphicx}

\makeatletter
\@namedef{ver@everyshi.sty}{}
\usepackage{tikz}
\makeatother
\usepackage{floatrow}
\usepackage{amsmath}
\usepackage{comment}

\DeclareMathOperator{\inv}{inv}
\DeclareMathOperator{\Inv}{Inv}
\DeclareMathOperator{\Exp}{Exp}
\DeclareMathOperator{\Log}{Log}
\DeclareMathOperator{\Mul}{Mul}
\DeclareMathOperator{\adj}{adj}
\DeclareMathOperator{\Adj}{Adj}
\DeclareMathOperator{\AdjT}{AdjT}
\DeclareMathOperator{\Act}{Act}
\DeclareMathOperator{\ActP}{ActP}

\usepackage[utf8]{inputenc}
\usepackage{listings}
\usepackage{xcolor}

\definecolor{codegreen}{rgb}{0,0.6,0}
\definecolor{codegray}{rgb}{0.5,0.5,0.5}
\definecolor{codepurple}{rgb}{0.58,0,0.82}
\definecolor{backcolour}{rgb}{0.95,0.95,0.92}

\lstdefinestyle{mystyle}{
    backgroundcolor=\color{backcolour},   
    commentstyle=\color{codegreen},
    keywordstyle=\color{magenta},
    numberstyle=\tiny\color{codegray},
    stringstyle=\color{codepurple},
    basicstyle=\ttfamily\footnotesize,
    breakatwhitespace=false,         
    breaklines=true,                 
    captionpos=b,                    
    keepspaces=true,                 
    numbers=left,                    
    numbersep=5pt,                  
    showspaces=false,                
    showstringspaces=false,
    showtabs=false,                  
    tabsize=2
}
\usepackage[pagebackref=true,breaklinks=true,colorlinks,bookmarks=false]{hyperref}

\def\cvprPaperID{****} 
\def\confYear{CVPR 2021}

\begin{document}

\title{Tangent Space Backpropagation for 3D Transformation Groups}

\author{Zachary Teed \ \ and \ \ Jia Deng \\
Princeton University \\
{\tt\small \{zteed,jiadeng\}@cs.princeton.edu}}

\maketitle

\begin{abstract}
We address the problem of performing backpropagation for computation graphs involving 3D transformation groups $SO(3)$, $SE(3)$, and $Sim(3)$. 3D transformation groups are widely used in 3D vision and robotics, but they do not form vector spaces and instead lie on smooth manifolds. The standard backpropagation approach, which embeds 3D transformations in Euclidean spaces, suffers from numerical difficulties. We introduce a new library, which exploits the group structure of 3D transformations and performs backpropagation in the tangent spaces of manifolds. We show that our approach is numerically more stable,  easier to implement, and beneficial to a diverse set of tasks. Our plug-and-play PyTorch library is available at \url{https://github.com/princeton-vl/lietorch}
\end{abstract}

\section{Introduction}

3D transformation groups---$SO(3)$, $SE(3)$, $Sim(3)$---have been extensively used in a wide range of computer vision and robotics problems. Important applications include SLAM, 6-dof pose estimation, multiview reconstruction, inverse kinematics, pose graph optimization, geometric registration, and scene flow. In these domains, the state of the system---configuration of robotic joints, camera poses, non-rigid deformations---can be naturally represented as 3D transformations.

Recently, many of these problems have been approached using deep learning, either in an end-to-end manner\cite{banet,deepglobalregistration,deeptam,se3net,densefusion,drisf,gradslam} or composed as hybrid systems\cite{learningrigidity,deepfactors,superglue,luo2020consistent}. A key ingredient of deep learning is auto-differentiation, in particular, backpropagation through a computation graph. A variety of general-purpose deep learning libraries such as PyTorch~\cite{pytorch} and TensorFlow~\cite{tensorflow} have been developed such that backpropagation is automatically performed by the library---users only need to specify the computation graph and supply any custom operations. 

A basic assumption of existing deep learning libraries is that a computation graph represents a composition of mappings between Euclidean spaces. That is, each node of the graph represents a differentiable mappings between Euclidean spaces, e.g.\@ from $\mathbb{R}^m$ to $\mathbb{R}^n$. This assumption allows us to use the standard definition of the gradient of a function $f(x): \mathbb{R}^n\rightarrow \mathbb{R}$ as $\frac{\partial f}{\partial x} = [\frac{\partial f}{\partial x_1}, \frac{\partial f}{\partial x_2}, \ldots,  \frac{\partial f}{\partial x_n}]$, and use the chain rule to backpropagate the  gradient: $\frac{\partial f}{\partial y} = \frac{\partial f}{\partial x}\cdot \frac{\partial x}{\partial y}$, where $\frac{\partial x}{\partial y}$ is the Jacobian matrix $(\frac{\partial x_i}{\partial y_j})$.

However, this assumption breaks down when the computation graph involves 3D transformations, such as when a network iteratively updates its estimate of the camera pose~\cite{banet,deeptam,deepv2d,deepglobalregistration,inversecomp}.  3D transformations do not form vector spaces and instead lie on smooth manifolds; the notion of addition is undefined and the standard notion of gradient defined in Euclidean spaces does not apply. 

A typical approach in prior work is to embed the 3D transforms in a Euclidean space. For example, a rigid body transform from $SE(3)$ is represented as a $4\times 4$ matrix and treated as a vector in $\mathbf{R}^{16}$. That is, $SE(3)$ corresponds to a subset of $R^{16}$, or more specifically, an embedded submanifold of $R^{16}$. Functions involving $SE(3)$ such as the exponential map and the inverse are replaced with their extensions in the embedding space, i.e.\@ the matrix exponential and the matrix inverse: 
\begin{equation}
\begin{aligned}
&\exp: \mathfrak{se}(3) \mapsto SE(3) \longrightarrow \overline{\exp}: \mathbb{R}^{6} \mapsto \mathbb{R}^{4\times4} \\
&\inv: SE(3) \mapsto SE(3) \longrightarrow \overline{\inv}: \mathbb{R}^{4\times4}  \mapsto \mathbb{R}^{4\times4}.
\end{aligned}
\end{equation}
Now, the computation graph involves only Euclidean objects, and backpropagation can be performed as usual as long as the Jacobian of $\overline{\exp}$ and $\overline{\inv}$ can be calculated. 

\begin{figure}[t]
    \centering
    \includegraphics[width=.9\columnwidth]{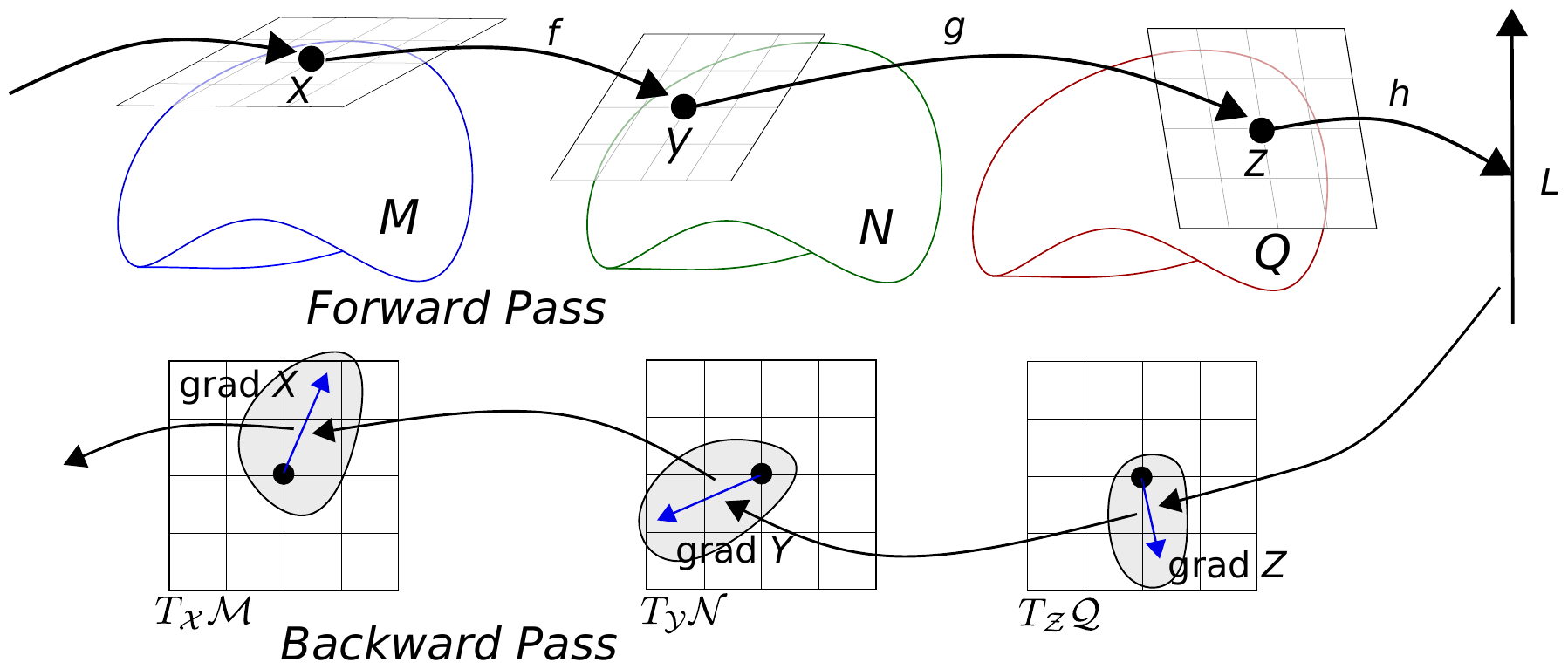}
    \caption{Tangent space Backpropagation. The forward pass is a composition of mappings between Lie groups. The backward pass propogates gradients in the tangent space of each element.}
    \label{fig:pull}
\end{figure}

There are several problems with this approach. First, while the extended functions such as the matrix exponential are smooth as a whole, the individual substeps needed for computing the backward passes often contain singularities. As a result, small deviations off the manifold can result in numerical instabilities causing gradients to explode. It can often be quite difficult to implement these functions in libraries such as PyTorch\cite{pytorch} and Tensorflow\cite{tensorflow} in a way that numerically stable gradients are achieved through automatic differentiation. As a case in point, the commonly used PyTorch3D library\cite{ravi2020accelerating} returns nan-gradients when the identity matrix is given as input the matrix log. Furthermore, some extended functions have very complicated backward pass, which leads to extremely large computation graphs, especially for groups like $Sim(3)$ which involve complex expressions for the matrix exponential and logarithm. Partly for this reason, to our knowledge, no prior work has performed backpropagation on $Sim(3)$. 

Due to the aforementioned problems, existing deep learning libraries are unable to handle 3D transformations reliably and transparently. And incorporating 3D transformations into a computation graph  typically requires a substantial amount of painstaking, ad-hoc manual effort. 

In this work, we seek to make backpropagation on 3D transformations robust and ``painless''. We introduce a new approach for performing backpropagation though mixed computation graphs consisting of both real vectors and 3D transformations. Instead of embedding transformation groups in Euclidean spaces, we retain the group structure and perform backpropagation  directly in the \emph{tangent space} of each group element (Fig.~\ref{fig:pull}). For example, while a rigid body transformation $T \in SE(3)$ may be represented as a $\mathbb{R}^{4 \times 4}$ matrix, we backpropagate the gradient in the tangent space $\mathfrak{se}(3)$, in particular, as a 6-dimensional vector in a local coordinate system centered at $T$.

We show that performing differentiation in the tangent space has several advantages
\begin{itemize}
\item[--] \emph{Numerical Stability:} By performing backpropagation in the tangent space, we avoid needing to differentiate through singularity-ridden substeps required for the embedding approach, guaranteeing numerically stable gradients. This allows us to provide groups such as $Sim(3)$ which do not give stable gradients when automatic differentiation is directly applied.
\item[--] \emph{Representation Agnostic:} The computation of the gradients does not depend on how the 3D transformations are represented. Both quaternions and $3\times3$ matrices can be used to represent rotations without changing how the backward pass is computed.
\item[--] \emph{Reduced Computation Graphs:} When functions such as \texttt{exp} and \texttt{log} are implemented directly in PyTorch, the output of each individual steps within \texttt{exp} and \texttt{log} need to be stored for the backward pass, leading to unnecessarily large computation graphs. We avoid the need to store intermediate values by differentiating through these functions using the group structure.
\item[--] \emph{Manifold Optimization:} Our library can be directly used for problems where the variables we want to optimize are 3D transforms. Since we compute gradients directly in the tangent space, we avoid the need to reproject gradients.  
\end{itemize}

We demonstrate use cases of our approach on a wide range of vision and robotics tasks. We show our approach can be used for pose graph optimization, inverse kinematics, RGB-D scan registration, and RGB-D SLAM. 

Our approach is implemented as an easy-to-use, plug-and-play PyTorch library. It allows users to insert 3D transformations, either as parameters or activations, into a computation graph, just as regular tensors; backpropagation is taken care of transparently. Our 3D transformation objects expose an interface similar to the \texttt{Tensor} object, supporting arbitrary batch shapes, indexing, and reshaping operations.

Our contributions are two-fold. First, we introduce a new method of auto-differentiation involving 3D transformation groups by performing backpropagation in the tangent space. To our knowledge, this is the first time backpropagation is performed in the tangent space of Lie groups for training neural networks. Second, we introduce LieTorch, an open-source, easy-to-use PyTorch library that implements tangent space backpropagation. We expect our library to be a useful tool for researchers in 3D vision and robotics.

\section{Related Work}

\vspace{1mm} \noindent \textbf{Automatic Differentiation:} Automatic Differentiation (AD) is a family of algorithms for evaluating derivatives of a program. AD frameworks expose a set of elementary operators (e.g. matrix multiplication, convolution, and pooling) and programs can be constructed by composing operations. Modern implementations can handle complex computations graphs with branching, loops, and recursion\cite{baydin2017automatic,pytorch}. 

AD has two common forms: \emph{forward mode differentiation} applies the chain rule to each elementary operator in the forward pass. Optimization frameworks such as Ceres\cite{ceres} implement forward mode differentiation using dual numbers. \emph{Reverse mode differentiation} is a general form of backpropagation and works by complementing each intermediate value with an gradient. During the forward pass, the intermediate values are populated. In the backward pass, gradients are propagated in reverse. Reverse mode differentiation is well suited for differentiating functions with a single objective $f: \mathbb{R}^n \mapsto \mathbb{R}$, such as the loss used to train a neural network. Deep learning frameworks such as Theano\cite{team2016theano}, Tensorflow\cite{tensorflow}, Autograd\cite{maclaurin2015autograd}, PyTorch\cite{pytorch}, and JAX\cite{jax2018github} all support reverse mode differentation.

Existing frameworks do not directly support manifold elements in the computation graph. These libraries assume every variable belongs to an Euclidean space and every function maps from one Euclidean space to another. However, 3D transformations groups such as rotations do not form a vector space, and the usual notions of derivatives do not apply. We use a more general notation of gradient defined in tangent spaces and show that we can support 3D transformation groups by performing differentiation in the tangent space. We build our library on top of PyTorch, and expose an interface similar to the \texttt{Tensor} object, supporting arbitrary batch shapes, indexing, and reshaping operations. By building on PyTorch, we can compose Lie group and tensor operations in a shared computation graph.

Numeric issues often arise in AD if operations are naively implemented.  For example, expressions such as the L2 norm $||\cdot||_2$ or the \texttt{LogSumExp} are problematic if implemented directly, and various tricks are required in order to ensure numeric stability. The exponential and logarithmic maps for 3D transformation groups contain many problematic expressions. We implement group operations (e.g. \texttt{exp}, \texttt{log}, \texttt{act}, \texttt{adj}) as the elementary operations.

\vspace{1mm} \noindent \textbf{Manifold Optimization:} Many vision and robotics problems requiring optimizing variables which lie on a manifold. Libraries such as GTSAM\cite{gtsam} and g2o\cite{g2o} provide general frameworks for solving nonlinear least-squares and MAP inference problems involving manifold elements such as camera poses. GTSAM and Koppel et al.\cite{koppel2018manifold} provide frameworks which can perform automatic differentiation over lie groups. However, these frameworks are tailored to the computation of Jacobian matrices and cannot be readily used within the computation graphs for training neural networks. 



There are several libraries which provide tools for optimization on manifolds, such as Manopt\cite{manopt}, and PyManopt includes autodifferentiation capabilities provided by PyTorch\cite{pytorch}, Tensorflow\cite{tensorflow}, and Autograd \cite{maclaurin2015autograd}. Several extensions have been proposed to PyTorch which allow optimization over smooth manifolds such as McTorch\cite{meghwanshi2018mctorch} and Geopt\cite{geoopt2020kochurov}. These libraries work by embedding manifolds in the Euclidean space $\mathbb{R}^n$, and manifold functions $f: \mathcal{N} \mapsto \mathcal{M}$ are implemented as the extension $\overline{f}: \mathbb{R}^n \mapsto \mathbb{R}^m$. Automatic differentiation can be used to differentiate $\overline{f}$, and the gradient on $\mathcal{M}$ is obtained using the orthogonal projection\cite{boumal2020introduction}.

This strategy can be prone to large computation graphs and numerical instabilities. Our library avoids the need to differentiate $\overline{f}: \mathbb{R}^n \mapsto \mathbb{R}^m$ by differentiating the original function $f: \mathcal{N} \mapsto \mathcal{M}$ directly, i.e.\@ obtaining the differential of $f$, which is a map from the tangent space of $\mathcal{N}$ to the tangent space of $\mathcal{M}$. We note that this is not possible for all manifolds, but using the additional structure provided by Lie Groups, we demonstrate that AD can be performed directly in the tangent space. An additional advantage of our approach is that we never need to perform a projection step, since gradients are already defined in the tangent space.

\section{Preliminaries}

A matrix Lie group $\mathcal{M}$ is both a group and a smooth manifold. Each element $X \in \mathcal{M}$ can be represented as a matrix in $\mathbb{R}^{n \times n}$. The group operator is identical to matrix multiplication and the group inverse is identical to matrix inversion. Being a smooth manifold, each element $X \in \mathcal{M}$ has a unique tangent space. Moreover, the tangent space of each group element is a vector space isomorphic to the tangent space of the identity element. 

\vspace{1mm} \noindent \textbf{Lie Algebra:} The lie algebra $\mathfrak{g}$ is defined as the tangent space at the identity element, and each group has an associated lie algebra. The lie algebra $\mathfrak{g}$ forms a vector space with a set of basis elements $\{ G_1, \hdots, G_k \}$. The lie algebra is isomorphic to $\mathbb{R}^k$, and we can map between elements of $\mathfrak{g}$ and elements of $\mathbb{R}^k$ using the hat $^\wedge$
\begin{equation}
    ^\wedge: \mathbb{R}^k \rightarrow \mathfrak{g}: \qquad \boldsymbol{\tau}^{\wedge} = \sum_i^k \tau_i G_i
\end{equation}
and the vee operator $^\vee: \mathfrak{g} \mapsto \mathbb{R}^k$ which is the inverse of the wedge operator, such that $\left(\boldsymbol{\tau}^\wedge \right)^\vee = \boldsymbol{\tau}$.
The lie algebra $\mathbf{g}$ is isomorphic to $\mathbb{R}^k$, $\mathfrak{g} \cong \mathbb{R}^k$. Since it is often easier to work in $\mathbb{R}^k$, we perform differentiating using vectors in $\mathbb{R}^k$, but it would be equivalent to representing gradients in $\mathfrak{g}$. The definitions of the $\wedge$ and $\vee$ operators for 3D transformation groups are given in the appendix. 

\vspace{1mm} \noindent \textbf{Exponential and Logarithm Map:} Elements of the lie algebra $\boldsymbol{\phi}^\wedge \in \mathfrak{g}$ can be exactly mapped to the manifold through the exponential map.
\begin{equation}
    \exp(\boldsymbol{\phi}^\wedge) = \mathbf{I} + \boldsymbol{\phi}^\wedge + \frac{1}{2!}(\boldsymbol{\phi}^\wedge)^2 + \frac{1}{3!}(\boldsymbol{\phi}^\wedge)^3 + \hdots
\end{equation}
The logarithm map is the inverse of the exponential map and takes elements from the manifold to the lie algebra. For convenience, we use vectorized versions of the exponential and logarithm map which map directly between $\mathbb{R}^k$ and $\mathcal M$
\begin{align}
    \Exp: \mathbb{R}^k \rightarrow \mathcal{M} \qquad \Log: \mathcal{M} \rightarrow \mathbb{R}^k   
\end{align}

\vspace{1mm} \noindent \textbf{Group Multiplication:} The group is endowed with a binary operator $\circ$ such that two group elements can be combined to form a third element $X \circ Y \in \mathcal{M}$. Using the notation of Sola et al. \cite{sola2018micro}, we can overload the the addition and subtraction operations
\begin{align}
    &\oplus: \mathbb{R}^k \times \mathcal{M} \rightarrow \mathcal{M} \qquad \xi \oplus X = \Exp(\xi) \circ X \\
    &\ominus: \mathcal{M} \times \mathcal{M} \rightarrow \mathbb{R}^k \qquad X \ominus Y  = \Log(X \cdot  Y^{-1})
\end{align}
If the manifold $\mathcal{M}$ is a Euclidean space, then $\oplus$ and $\ominus$ are identical to standard vector addition and subtraction.

\vspace{1mm} \noindent \textbf{Adjoint:} The Lie algebra and exponential map give us two possible local parameterizations of a neighborhood around a group element $X$: the ``right action parameterization'' $X \circ \Exp(a)$ and the ``left action parameterization'' $\Exp(b) \circ X$, where $a$ and $b$ represent the local coordinates. For a group element around $X$, its coordinates $a$ and $b$ are related by a linear map, which is the adjoint of $X$:

\begin{equation}
    \Adj_X(a) = (X a^\wedge X^{-1})^\vee, 
\end{equation}
from which it is easy to verify that 
\begin{equation}
     X \circ \Exp(a) = \Exp(\Adj_X(a)) \circ X. 
     \label{eqn:adjoint}
\end{equation}
Because $Adj_X$ is a linear map,  we $\mathbf{Adj}_X$ to denote its matrix representation, i.e.\@ $\mathbf{Adj}_X$ is a $n\times n$ matrix where $n$ is the dimension of the tangent space of $X$. The expressions of the adjoint for 3D transformation groups are given in the appendix.

\begin{table}[t]
\footnotesize
\centering
\begin{tabular}{lll}
\toprule
Operation & Map & Description \\
\midrule
$\Exp$ & $\mathfrak{g} \mapsto \mathcal{M}$ & exponential map \\
$\Log$ & $\mathcal{M} \mapsto \mathfrak{g}$ & logarithm map \\
$\Inv$ & $\mathcal{M} \mapsto \mathcal{M}$ & group inverse \\
$\Mul$ & $\mathcal{M} \times \mathcal{M} \mapsto \mathcal{M}$& group multiplication \\
$\Adj$ & $\mathcal{M} \times \mathfrak{g} \mapsto \mathfrak{g}$ & adjoint operator \\
$\AdjT$ & $\mathcal{M} \times \mathfrak{g}^* \mapsto \mathfrak{g}^*$ & dual adjoint operator \\
$\Act$ & $\mathcal{M} \times \mathbb{R}^3 \mapsto \mathbb{R}^3$ & action on point (set) \\
$\ActP$ &  $\mathcal{M} \times \mathbb{P}^3 \mapsto \mathbb{P}^3$ & action on homogeneous point (set) \\
\bottomrule
\end{tabular}
\caption{Summary of operations supported by our library. Each operation is differentiable with respect to all the input arguments. Both the lie algebra $\mathbf{g}$ and its dual $\mathbf{g}^*$ are embedded in $\mathbb{R}^k$.}
\label{table:Groups}
\end{table}

\vspace{1mm} \noindent \textbf{3D Transformation Groups} In this paper we are particularly concerned a special class of Lie groups that perform 3D transformation---SO(3), SE(3), and Sim(3). A description of the supported group operations is given in Tab.~\ref{table:Groups}.

\vspace{1mm} \noindent \textbf{Differentials and Gradients:} Given a function $f: \mathbb{R}^n \rightarrow \mathbb{R}^m$ and a point $x \in \mathbb{R}^n$ the differential of $f$ is the linear operator
\begin{equation}
    D f(x)[v] = \lim_{t \rightarrow 0} \frac{f(x + tv) - f(x)}{t}
\end{equation}
However, this is problematic for functions on Lie groups $f: \mathcal{M} \rightarrow \mathcal{M}'$ since in general $\mathcal{M}$ is not closed under addition. However, we can generalize the differential as perturbations in the tangent space
\begin{equation}
   D f(X)[\mathbf{v}] = \lim_{t\rightarrow 0} \frac{f(t\mathbf{v}\oplus X) \ominus f(X)}{t}
    \label{eqn:differential}
\end{equation}
where $v$ belongs to the tangent space of $X$. Eqn. \ref{eqn:differential} relates perturbations in the tangent space of $X$ to perturbations in the tangent space of $f(X)$.
Plugging in the (orthonormal) basis vectors into Eqn. \ref{eqn:differential} we can recover the Jacobian matrix $\mathbf{J} \in \mathbb{R}^{n \times m}$
\begin{equation}
   \mathbf{J}_{ij} = \lim_{t\rightarrow 0} \frac{<f(t\mathbf{e}_j\oplus X) \ominus f(X), \mathbf{e'}_i>}{t},
   \label{eqn:jacobian}
\end{equation}
where $<,>$ is an inner product defined on the tangent space of $\mathcal{M}'$. 

Throughout this paper we will use $\frac{\partial Y}{\partial X}$ to denote the Jacobian of a mapping $f: X \mapsto Y$ between two Lie groups, under the basis vectors given by the left action parameterizations of the tangent spaces of $Y$ and $X$. That is, the dimension of $\frac{\partial Y}{\partial X}$ is $m\times n$, where $m$ is the dimension of the tangent space of $Y$ and $n$ is the dimension of the tangent space of $X$. 

Note that under vector addition, a Euclidean space is a Lie group whose tangent space is itself, so this definition subsumes the standard notion of Jacobian. In particular, for a loss function $L(X) \in \mathbb{R}$, $\frac{\partial L}{\partial X}$ is a row vector representing the gradient of $L$ in the tangent space of $X$. 

\section{Approach}

\begin{figure}[t]
\tikzstyle{vertex}=[circle,fill=black!15,minimum size=30pt,inner sep=0pt]
\tikzstyle{selected vertex} = [Large,rectangle, fill=red!24]
\tikzstyle{edge} = [draw,thick,->]
\tikzstyle{edge1} = [draw,thick,<-]
\tikzstyle{weight} = [font=\small]
\tikzstyle{selected edge} = [draw,line width=5pt,-,red!50]
\tikzstyle{ignored edge} = [draw,line width=5pt,-,black!20]

\resizebox{1.0\textwidth}{!}{%
\begin{tikzpicture}
    \node[](in1) at (0, 0) {$\mathbf{G}_1$};
    \node[](in2) at (8, -2.2) {$\mathbf{T}$};

    \node[](odot1) at (2, 0) {$\odot$};
    \node[rectangle,fill=green!20](exp1) at (2,-1) {$\Exp$};

    \node[](odot2) at (4, 0) {$\odot$};
    \node[rectangle,fill=green!20](exp2) at (4,-1) {$\Exp$};

    \node[](odot3) at (6, 0) {$\odot$};
    \node[rectangle,fill=green!20](exp3) at (6,-1) {$\Exp$};
    
    \node[](odot4) at (8, 0) {$\odot$};
    \node[rectangle,fill=blue!20](log) at (10,0) {$\Log$};
    \node[rectangle,fill=red!20](norm) at (12,0) {$|| \cdot ||$};
    \node[rectangle,fill=yellow!20](inv) at (8,-1) {$\Inv$};

    \node[](f1) at (2,-2.2) {$A$};
    \node[](f2) at (4,-2.2) {$A$};
    \node[](f3) at (6,-2.2) {$A$};
    \node[rectangle,align=center,fill=gray!20,text width=60mm](network) at (4,-2.2) {$f_\theta$};

    \node[](loss) at (13.5, 0) {$\mathcal{L}$};
    \path[edge] (in1) -- node[weight,pos=0.5,above] {} (odot1) ;
    \path[edge] (odot1) -- node[weight,pos=0.5,above] {$\mathbf{G}_2$} (odot2) ;
    \path[edge] (odot2) -- node[weight,pos=0.5,above] {$\mathbf{G}_3$} (odot3) ;
    \path[edge] (odot3) -- node[weight,pos=0.5,above] {$\mathbf{G}_4$} (odot4) ;
    \path[edge] (odot4) -- node[weight,pos=0.5,above] {$\mathbf{E}$} (log) ;
    \path[edge] (log) -- node[weight,pos=0.5,above] {$\boldsymbol{\phi}$} (norm) ;
    \path[edge] (norm) -- node[weight,pos=0.5,above] {} (loss) ;
    
    \path[edge] (exp1) -- node[weight,pos=0.5,right] {$\mathbf{H}_1$} (odot1) ;
    \path[edge] (exp2) -- node[weight,pos=0.5,right] {$\mathbf{H}_2$} (odot2) ;
    \path[edge] (exp3) -- node[weight,pos=0.5,right] {$\mathbf{H}_3$} (odot3) ;
    
    \path[edge] (in2) -- node[weight,pos=0.5,above] {} (inv) ;
    \path[edge] (inv) -- node[weight,pos=0.5,above] {} (odot4) ;
    
    \path[edge] (f1) -- node[weight,pos=0.5,right] {$\delta_1$} (exp1) ;
    \path[edge] (f2) -- node[weight,pos=0.5,right] {$\delta_2$} (exp2) ;
    \path[edge] (f3) -- node[weight,pos=0.5,right] {$\delta_3$} (exp3) ;
\end{tikzpicture}
}

\caption{Computation graph involving Lie groups. A network $f_\theta$ producs a series of updates $\delta_1, \delta_2, \delta_2$ which are applied to pose $\mathbf{G}_1$. The loss function is defined on the geodesic distance between the estimated pose and the ground truth pose $\mathbf{T}$.}
\label{fig:example_graph}
\end{figure}

Our approach performs backpropagation through computation graphs consisting of Lie group elements and operations which map between groups. As an example, consider the computation graph shown in Fig. \ref{fig:example_graph}. The computation graph in Fig. \ref{fig:example_graph} can be written as
\begin{equation}
    \mathcal{L} = ||\mathbf{T}^{-1} \circ (e^{\delta_1}e^{\delta_2}e^{\delta_3}\mathbf{G}_1) ||
\end{equation}
where a series of updates $\delta_1, \delta_2, \delta_3$ are predicted by a network $f_\theta$. The loss is then taken to be the norm of the geodesic distance between $e^{\delta_1}e^{\delta_2}e^{\delta_3}\mathbf{G}_1$ and the ground truth pose $\mathbf{T}$. This type of computation graph shows up in many different applications where incremental updates are applied to an initial estimate.

\subsection{Reverse Mode Autodifferentation}
Each node in the computation graph (Fig. \ref{fig:example_graph}) represents an element $X$ on some Lie group $\mathcal{M}$ and each edge in the computation graph represents a function which maps from one Lie group $\mathcal{N}$ to another $\mathcal{M}$. 
\begin{equation}
    f: \mathcal{M} \rightarrow \mathcal{N}, \qquad X \in \mathcal{M}, \ f(X) \in \mathcal{N}
    \label{eqn:manifold_function}
\end{equation}
This notation subsumes standard Euclidean elements such as vectors and matrices, which also form a Lie groups under vector addition.

During the forward pass, we evaluate each function in the computation graph in topological order. The output of each function is stored for the backward pass. The backward pass propagates gradients in reverse topological order. Given a function as defined in Eqn. \ref{eqn:manifold_function}, $Y = f(X)$, we can backpropagate the gradient using the chain rule

\begin{equation}
    \frac{\partial \mathcal{L}}{\partial X} = \frac{\partial \mathcal{L}}{\partial Y} \frac{\partial Y}{\partial X} =  \frac{\partial \mathcal{L}}{\partial Y} \mathbf{J}
    \label{eqn:chain}
\end{equation}
where $\frac{\partial \mathcal{L}}{\partial X}$ is a row vector with the same dimension as the tangent space of $X$ and $\mathbf{J}$ is the Jacobian in Eqn \ref{eqn:jacobian}. 
Eqn. \ref{eqn:chain} computes a Jacobian-vector product.

\subsection{Computing the Jacobians}
We use Eqn. \ref{eqn:differential} to derive analytical expressions for the Jacobians. As an example, we show how the Jacobians can be computed for two functions: group multiplication and the logarithm map. The derivations for other functions including the group inverse and the exponential map are included in the appendix. 

\vspace{1mm} \noindent \textbf{Group Multiplication:} Consider the first group multiplication in Fig. \ref{fig:example_graph}: $Z = X \circ Y$. Using the definition of the differential (Eqn. \ref{eqn:differential}), we first compute the differential with respect to $X$
\begin{align}
    D f(X)[\mathbf{v}] &=\lim_{t\rightarrow 0} \frac{\Log((e^{t\mathbf{v}}XY)(XY)^{-1})}{t} \\
    & = \lim_{t\rightarrow 0} \frac{\Log(e^{t\mathbf{v}})}{t} = \lim_{t\rightarrow 0} \frac{t \mathbf{v}}{t} = \mathbf{v}
\end{align}
The differential with respect to $\mathcal Y$ can be derived in a similiar manner
\begin{align}
    D f(Y)[\mathbf{v}] = \lim_{t\rightarrow 0} \frac{\Log((X e^{t\mathbf{v}} Y)(X Y)^{-1})}{t} \\
\end{align}
applying the adjoint (Eqn. \ref{eqn:adjoint})
\begin{align}
    &=\lim_{t\rightarrow 0} \frac{\Log(e^{\mathbf{Adj}_{X}\cdot t\mathbf{v}}(X Y)(X Y)^{-1})}{t} \\
    &= \lim_{t\rightarrow 0} \frac{\Log(e^{\mathbf{Adj}_{X}\cdot t\mathbf{v}})}{t} = \mathbf{Adj}_{X}\cdot\mathbf{v}
\end{align}
Using equation Eqn, \ref{eqn:chain}, we can propogate the gradients as
\begin{equation}
    \frac{\partial \mathcal L}{\partial X} = \frac{\partial \mathcal L}{\partial Z} \qquad 
    \frac{\partial \mathcal L}{\partial Y} = \frac{\partial \mathcal L}{\partial Z} \mathbf{Adj}_{X}.
\end{equation}
As an example, for $R \in SO(3)$, $\mathbf{Adj}_{R} = R$. 

\vspace{1mm} \noindent \textbf{Logarithm Map:} The logarithm map $\phi = \Log(X)$ takes a group element to its Lie algebra. As example, for $SO(3)$ its logarithm map $\Log: SO(3) \rightarrow \mathbb{R}^3$ can be expressed as 
\begin{equation}
    \Log(X) = \frac{\psi (X - X^T)^\vee}{2\sin(\psi)}, \psi = \cos^{-1}\left(\frac{tr(X)-1}{2}\right)
    \label{eqn:so3_log}
\end{equation}
Using Eqn. \ref{eqn:differential}, we can express the differential of the logarithm map (between tangent spaces) as
\begin{equation}
    D \Log(X)[\mathbf{v}] = \lim_{t\rightarrow 0} \frac{\Log(\Exp(t\mathbf{v})\circ X) - \Log(X)}{t}
    \label{eqn:differential_log}
\end{equation}
From the Baker-Campbell-Hausdorff formula~\cite{barfoot2017state} we have
\begin{equation*}
    \log(\exp(\delta \xi) \exp(\phi)) \approx \phi + \mathbf{J}_l^{-1}(\phi) \delta \phi, \mbox{ when } \delta \phi \mbox{ is small},
\end{equation*}
where $\mathbf{J}_l(\phi)$ is a matrix called ``the left Jacobian''~\cite{sola2018micro} which maps a perturbation in the tangent space to a perturbations on the manifold.
As an example, $\mathbf{J}^{-1}_l$ of $SO(3)$ has a closed form expression \cite{sola2018micro}:
\begin{equation}
    \mathbf{J}^{-1}_l (\phi) = \mathbf{I}_{3\times 3} - \frac{1}{2} \phi^\wedge + \left( \frac{1}{\phi^2} - \frac{1+\cos\phi}{2\phi \sin\phi} \right) (\phi^\wedge)^2
    \label{eqn:inverse_jacobian_so3}
\end{equation}
We can then derive the backpropagated gradient as (full derivation in the appendix): 
\begin{equation}
    \frac{\partial \mathcal L}{\partial X} = \frac{\partial \mathcal L}{\partial \phi}  \cdot \mathbf{J}_l^{-1}(\phi),
\end{equation}
which is a vector in the tangent space. For SO(3), the gradient is 3-dimensional.  

There also exists a closed form for $\mathbf{J}_l^{-1}$ for the $SE(3)$ group. For Lie groups such as $Sim(3)$ without an analytic expression for the left Jacobian, we can numerically approximate the gradient (to some desired precision) using the series expansion \cite{barfoot2017state}
\begin{equation}
    \mathbf{J}_l^{-1}(\phi) = \sum_{n=0} {(-1)}^n\frac{B_n}{n!} {(\phi^\curlywedge)}^n.
\end{equation}
where $\phi^\curlywedge = \mathbf{adj}(\phi^\wedge)$ and $\mathbf{adj}$ is the adjoint of the Lie algebra $\mathfrak{sim}(3)$. $B_n$ are the Bernoulli numbers. For small $\phi$, we can numerically approximate $\mathbf{J}_l(\phi) \approx \mathbf{I}$.

\vspace{1mm} \noindent \textbf{Embedding Space vs. Tangent Space} The logarithm map in $SO(3)$ is a good example to show the advantages of our approach over the standard embedding space backpropagation (e.g.\@ Autograd in Pytorch), which simply auto-differentiates the expressions given in Eqn.~\ref{eqn:so3_log} and obtains a 9-dimensional gradient, as opposed to a 3-dimensional gradient in our approach. 

The forward pass is the same for our approach and the standard approach---both use Eqn.~\ref{eqn:so3_log}. Eqn.~\ref{eqn:so3_log} as a whole is smooth but contains numerically problematic terms such as $\frac{\psi}{\sin\psi}$.  The solution is to use approximations given by its Taylor expansion $\frac{x}{\sin\psi} \approx 1 + \frac{1}{6} \psi^2 + \ldots$, when $\psi$ is small. This replacement is done only around singular points, corresponding to a dynamic modification of the computation graph during the forward pass. 

The backward pass, however, causes two difficulties for the standard backpropagation. First, it needs to handle backpropagating through numerically unstable terms such as  $\frac{\psi}{\sin\psi}$. In existing work, this is done by simply backpropagating through the Taylor approximation formula. However, the gradient of the Taylor approximation is not necessarily a good approximation of the true gradient. As a result, when to use and how many terms to use need to be very carefully tuned. For example, it is common to have the term $\frac{1}{x^2}$ in a Taylor approximation; including it risks division by zero in the term $\frac{1}{x^3}$ in the backward pass, but excluding it risks deviation from the manifold in the forward pass. 

Second, the standard backpropagation needs to handle terms with singular gradients. In Eqn.~\ref{eqn:so3_log}, the term $\cos^{-1}\left(\frac{tr(X)-1}{2}\right)$ has a singular gradient when $X$ is identity because $\frac {\partial}{\partial x} \cos^{-1}(x)$ is undefined at $x=1$. This issue in fact remains unaddressed in existing libraries. For example,  PyTorch3D~\cite{ravi2020accelerating} returns a NaN gradient for its matrix logarithm when the identity matrix is given as input.

In contrast, our approach suffers from none of these difficulties. For the backward pass we simply use Eqn.~\ref{eqn:inverse_jacobian_so3}, which is as straightforward as computing the forward pass.

\subsection{Implementation}
One of the technical challenges is integrating our tangent space representation into automatic differentiation software. We implement our library as an extension to PyTorch. We define a new type to represent group elements, and subclass this type for different groups. The resulting computation graph consists of mixed types, both Euclidean vectors and group elements. We implement a custom gradient for any function with group inputs or outputs, including the exponential map, logarithm map, adjoint, group inverse, group multiplication, and action on a point set.

Our implementation is a plug-and-play extension on PyTorch. To enable general use cases, we support arbitrary batch shapes, and common tensor operations such as indexing, reshaping, slicing, and repeating dimensions. We use unit quaternions to represent rotations since they are compact and have desirable numeric properties. All operations involving groups include both Cuda and c++ kernels to leverage the GPU if available.

\section{Experiments}
We show that our library can help a wide range of tasks.

\smallskip \noindent \textbf{Inverse Kinematics}
We first evaluate our library on a toy inverse kinematics task. Given a robot arm with joint lengths $d_1,...,d_N \in \mathbb{R}^+$ and a target $x^*$, the task is to find a set of relative rotations $\Delta \mathbf{R}_1, \Delta \mathbf{R}_2, ..., \Delta \mathbf{R}_N$ such that the end of the arm is positioned at $x^*$. Given the relative joint angles, we can use forward kinematics to compute the arm position
\begin{align}
    &\mathbf{R}_i = \Delta \mathbf{R}_i \cdot \Delta \mathbf{R}_{-1} \cdot \cdots \cdot \Delta \mathbf{R}_1 \\
    &y = \sum_i^N \mathbf{R}_i \cdot \begin{pmatrix} d_i & 0 & 0 \end{pmatrix}^T, \ \ \mathcal{L} = ||y - x^*||_2^2
\end{align}
We also experiment with extendable joints. These can be represented as the group of rotation and scaling in 3D or the $\mathbb{R}^+ \times SO(3)$ group, which can be represented as a rotation $\mathbb{R} \in SO(3)$ and a scaling $s \in \mathbb{R}^+$, $s\mathbf{R} \in \mathbb{R}^+ \times SO(3)$

We compare our approach to two different PyTorch/Autograd implementations. First, we directly backpropagate through each group operator, using the implementions provided by our library. For small angles, Backpropagation is performed through a Taylor approximation of the functions.We also show the performance of Autograd when the operations are explictly tuned for better stability. This includes reimplementing normalization functions, decreasing the threshold when Taylor approximations are used, and making division operations safe for small angles by adding a small epsilon to the denominator.
\begin{table}[h]
		\centering
		\resizebox{.8\textwidth}{!}{
		\begin{tabular}{l c c }
        \textbf{} & $SO(3)$ & $\mathbb{R}^+ \times SO(3)$\\
        \midrule
        PyTorch+Autograd & 0.0 & 0.0 \\
        PyTorch+Autograd (tuned) & 99.8 & \textbf{100.0} \\
        \midrule
        Ours & \textbf{100.0} & \textbf{100.0} \\
        \bottomrule
        \end{tabular}
        }
	\label{table:kinematics}
\end{table}

The results from these experiments are provided in Tab. \ref{table:kinematics}. We perform 1000 runs for both the $SO(3)$ and $\mathbb{R}^+ \times SO(3)$ experiments. We report the portion of runs which converge to the correct solution within 1000 iterations using a tolerance threshold of $1\times 10^{-4}$.

We see that without tuning the forward pass, PyTorch+Autograd diverges on every single problem. By modifying the forward expression to make it safe for automatic differentation, we can get near 100\% convergence. Our library converges on all the problems without modification.

\begin{figure} [th]
    \centering
    \includegraphics[width=\columnwidth]{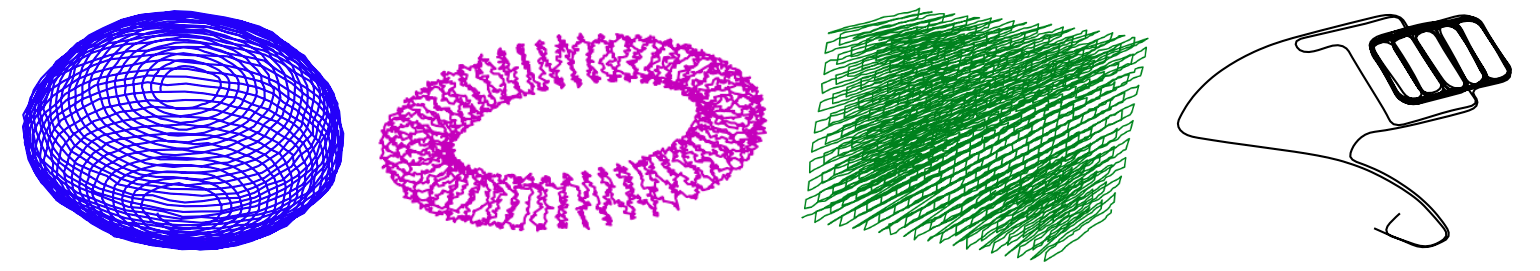}
	\caption{Optimized pose graphs on (left to right) sphere, torus, grid, and parking-garage problems. We perform gradient descent on the rotation group as initialization, followed by 7 Gauss-Newton updates.}	
	\label{fig:posegraph}
\end{figure}

\begin{table*}[th]
		\centering
		\resizebox{.8\textwidth}{!}{
		\begin{tabular}{l|c c c c | c c}
        \textbf{} & g2o\cite{g2o} & gtsam\cite{gtsam} & chordal+gtsam\cite{carloneinitialization} & gradient+gtsam\cite{carloneinitialization} & PyTorch+Autograd & Ours \\
        \midrule
        Parking-Garage & $6.42\times10^{-1}$ & $\mathbf{6.35\times10^{-1}}$ & $\mathbf{6.35\times10^{-1}}$ & $\mathbf{6.35\times10^{-1}}$ & $\mathbf{6.35\times10^{-1}}$ & $\mathbf{6.35\times10^{-1}}$ \\
        ($n=1661$, $m=6275$) & - & - & 0.23 & 1.29 & 16.5 & 1.18  \\

        \midrule
        Sphere-A & $5.32\times10^{10}$ & $5.71\times10^{10}$ & $\mathbf{1.49\times10^6}$ & $1.9\times10^6$ & $\mathbf{1.49\times10^6}$ & $\mathbf{1.49\times10^6}$ \\
        ($n=2200$, $m=8647$) & - & - & 0.88 & 46.3 & 16.9 & 1.19  \\
        \midrule
        Torus & $6.04\times10^{8}$ & $4.71\times10^{10}$ & $\mathbf{1.21\times10^4}$ & $2.81\times10^4$ & $\mathbf{1.21\times10^4}$ & $\mathbf{1.21\times10^4}$ \\
        ($n=5000$, $m=9048$) & - & - & 1.18 & 20.2 & 17.2 & 1.17  \\
        \midrule
        Cube & $5.39\times10^{7}$ & $6.59\times10^{11}$ & $\mathbf{4.22\times10^4}$ & $\mathbf{4.22\times10^4}$ & $\mathbf{4.22\times10^4}$ & $\mathbf{4.22\times10^4}$ \\
        ($n=8000$, $m=22236$) & - & - & 17.9 & 26.4 & 18.3 & 1.21 \\
        \bottomrule
        \end{tabular}
        }
	\label{table:pose_graph}
	\caption{Error (top row) and time (bottom row) on pose graph optimization. Time (seconds) is reported for the initialization method; the first two columns are run without any initiliation.}
\end{table*}

\smallskip \noindent \textbf{Pose Graph Optimization}
Pose graph optimization is the problem of recovering the trajectory of a robot given a set of noisy measurements. Conventional methods use iterative solvers such as Gauss-Newton\cite{gtsam} or Levenberg-Marquardt\cite{g2o}. However, due to rotation, pose graph optimization is a non-convex optimization problem, and second order methods are prone to local minimum \cite{carloneinitialization}. 

One solution for overcoming local minimum is to use a good initialization. \emph{Riemannian gradient descent}\cite{tron2014distributed} provides an initialization for rotation by performing gradient descent using a reshaped cost function. These rotations are then used as initialization for second order methods. 

We perform optimization over all rotations jointly, using the reshaped cost function over the geodesic distance using the reshaping function proposed by Tron et al. \cite{tron2012intrinsic}
\begin{align}
    &\theta = || \log(\mathbf{R}_i^{-1} \cdot \mathbf{R}_j \cdot \mathbf{R}_{ij}^{-1}) ||_2 \\
    &\mathcal{L}(\theta) = 1/b-(1/b+\theta)\exp(-b\theta)
\end{align}
where $\mathbf{R}_{ij}$ are the noisy measurements and we set $b=1.5$. For optimization we use SGD with momentum set to 0.5. We perform 1000 gradient steps and a exponentially decaying learning rate $.995^\gamma$. Riemannian gradient descent can be easily implemented using our library with only a few lines of code, as shown in the sample below

\begin{lstlisting}[language=Python]
# (ii, jj) edges in pose graph
from lietorch import SO3

Eij = SO3(torch.from_numpy(Eij)).to('cuda')
R = SO3(torch.from_numpy(R)).to('cuda'))
optimizer = optim.SGD([R], lr=1.0, momentum=0.5)

for i in range(n_steps):
    optimizer.zero_grad()
    dE = (R[ii].inv() * R[jj]) * Eij.inv()
    loss = reshaping_fn(dE)

    loss.backward()
    optimizer.step()
\end{lstlisting}

We follow the setup by Carlone et al.\cite{carloneinitialization} and report convergence and timing results in Tab.~\ref{table:pose_graph}; optimized pose graphs are shown in Fig.~\ref{fig:posegraph}. We compare to g2o\cite{g2o} and gtsam\cite{gtsam}; gtsam also provides implementations of different initialization strategies such as chordal relaxation\cite{martinec2007robust} and Riemannian gradient descent\cite{tron2012intrinsic}. We test both our approach and a default PyTorch/Autograd implementation where backpropagation is performed directly in the embedding space. 

On all datasets, we find that our gradient based initialization converges to the global minimum, matching the performance of chordal relaxation. On two datasets, the gradient based initializer in gtsam gets stuck in a local minimum, while our implementation is able to converge. On all datasets, our implementation is much faster than gradient+gtsam since our library can leverage the GPU. chordal+gtsam is faster on the smaller problems, but our gradient based initializer is much faster on larger problems. We find that our method converges to the same solution as Autograd, but since our library performs a much simpler backward pass, it can properly leverage the GPU, consistently providing a 10-15x speedup.

\begin{table*}
\centering
	\resizebox{.75\textwidth}{!}{
	\begin{tabular}{l c c | c c c }
	& \multicolumn{2}{c}{\underline{$SE(3)$}} & \multicolumn{3}{c}{\underline{$Sim(3)$}} \\
    & tr. ($<$1cm) & rot. ($<0.1^\circ$) & tr. ($<$1cm) & rot. ($<0.1^\circ$) & scale ($<$1\%) \\
    \midrule
    Identity & 0.05 & 0.05 & 0.05 & 0.05 & 0.65 \\
    PyTorch+Autograd & 0.0 (NaN) & 0.0 (NaN) & 0.0 (NaN) & 0.0 (NaN) & 0.0 (NaN) \\
    PyTorch+Autograd (tuned) & $78.75\pm0.25$ & $90.9\pm0.25$ & $77.0\pm0.25$ & $90.8\pm0.25$ & $98.0\pm0.25$ \\
    \midrule
    Ours (1st order) & $78.55\pm0.28$ & $91.1\pm0.40$ & $77.4\pm0.28$ & $90.9\pm0.40$ & $98.1\pm0.43$ \\
    Ours (2nd order) & $\mathbf{79.0} \pm0.59$ & $91.0\pm0.39$ & $77.3\pm0.59$ & $91.1\pm0.39$ & $\mathbf{98.3}\pm0.35$ \\
    Ours (3rd order) & $78.6\pm0.48$ & $\mathbf{91.2}\pm0.58$ & $\mathbf{77.4}\pm0.48$ &  $\mathbf{91.2}\pm0.58$ & $97.2\pm0.25$ \\
    Ours (Analytic) & $78.7\pm0.33$ &  $91.1\pm0.36$ & - & - & - \\
    \bottomrule
    \end{tabular}
    }
    \caption{Results on RGB-D registration task. We provide the mean and standard deviation over 3 runs each.}
\label{table:sim3}
\end{table*}

\begin{table*} [h]
\centering
\resizebox{.8\linewidth}{!}{%
\begin{tabular}{l|lllllllll | l}
& 360 & desk & desk2 & floor & plant & room & rpy & teddy & xyz & avg\\
\midrule
DeepTAM \cite{deeptam} & 0.111 & 0.053 & 0.103 & \textbf{0.206} & 0.064 & 0.239 & 0.093 & 0.144 & 0.036 & 0.116\\
DeepV2D\cite{deepv2d} & \textbf{0.072} & \textbf{0.069} & 0.074 & 0.317 & 0.046 & 0.213 & 0.082& \textbf{0.114} & \textbf{0.028} & 0.113 \\
DeepV2D (ours) & 0.096 & 0.077 & 
\textbf{0.072} & 0.268 & \textbf{0.024} & \textbf{0.173} & \textbf{0.057} & 0.136 & 0.040 & \textbf{0.105} \\
\midrule
\end{tabular}
}
\caption{Tracking results in the RGB-D benchmark (ATE rmse [m]).}
\label{table:SLAM}
\end{table*}

\smallskip \noindent \textbf{RGB-D Sim3 Registration}
Given two RGB-D scans captured from different poses, we want to find a similarity transformation which aligns the two scans. This registration problem shows up in scan-to-CAD registration and loop closure in monocular SLAM\cite{mur2015orb}. While several recent works have used deep networks for registration \cite{deepglobalregistration,gojcic2020learning}, they have focused on recovering a $SE(3)$ transformation. Here, we demonstrate that our library can recover a $Sim(3)$ transformation that allows scale change.

\begin{figure}[t]
    \centering
    \includegraphics[width=\columnwidth]{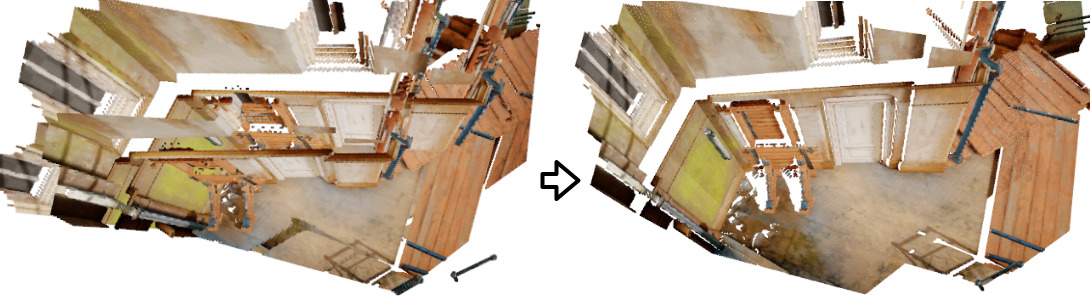}
	\caption{Example Sim3 registration, the network takes two RGB-D scans as input and predicts a similiarity transformation which aligns the two scans.}
	\label{fig:sim3registration}
\end{figure}

We run our experiments on the synthetic TartanAir dataset\cite{wang2020tartanair}, taking the scenes \emph{westerndesert, seasidetown, seasonsforest\_winter, office2} and \emph{gascola} for testing and the remaining scenes for validation and training. The network is trained to predict a $SE(3)$ or $Sim(3)$ transformation between a pair of frames.

In order to select training pairs which have sufficient difficulty, we compute the magnitude of the optical flow between all pairs of frames and uniformly sample pairs where the mean flow magnitude is between 16 and 120 pixels, resulting in a median translation of 54cm and a median rotation angle of 4.7$^\circ$.. For Sim(3), we sample scaling uniformly in the range $[0.5, 2.0]$ and rescale the depth maps accordingly.

We use a network architecture based on RAFT\cite{raft} and build a full 4D correlation volume by computing the visual similarity between all-pairs of pixels in the input images (more details on the network architecture are provided in the appendix). We start by initializing the estimate of the 3D similarity transform to be the identity element $\mathbf{I} \in Sim(3)$. The network predicts a series of updates $\Delta \mathbf{T}_k \in Sim(3)$ which are applied to the current estimate. 

In each iteration, we use the current estimate of the transformation to compute the optical flow and inverse depth change. We use the optical flow to index from the correlation volume similiar to RAFT. A GRU uses the correlation features and inverse depth error to output pixelwise residuals $r_x, r_y, r_z \in \mathbb{R}$ along with respective confidence weights $w_x, w_y, w_z \in [0,1]$. $r_x$ and $r_y$ is the residual flow in the $x$ and $y$ directions, while $r_z$ is the residual inverse depth.

The residual terms are used as input to a differentiable least squares optimization layer, which performs 3 Gauss-Newton updates which update the estimated transformation through $T_{k+1} = \Delta x \oplus T_k$. During training, we use our library to backprop through the Gauss-Newton updates to train the weights of the network. We train the network using the geodesic loss
\begin{equation}
    \vspace{-2mm}
    \mathcal{L}(\mathbf{T}_1, \hdots, \mathbf{T}_K) = \sum_k || \Log(\mathbf{T}_k^{-1} \cdot \mathbf{T}^*) ||
    \vspace{-2mm}
    \label{eq:geodesic}
\end{equation}
where $\mathbf{T}^*$ is the ground truth transformation.

We provide results from this experiment in Tab. \ref{table:sim3}, and qualitative results are shown in Fig. \ref{fig:sim3registration}. As we note earlier, there isn't a closed form expression for the left Jacobian of $Sim(3)$. Hence, we use a series approximation based on the Taylor expansion. We find that the order of the approximation makes little difference in the final accuracy. To the best of our knowledge, this is the first time backpropagation has been performed on similarity transformations, which is easily enabled by our approach.

\smallskip \noindent \textbf{RGB-D SLAM}
In our final experiment, we use our library to implement an deep RGB-D SLAM system. We reimplement the approach from DeepV2D\cite{deepv2d} in PyTorch so that it can be directly used with our library. DeepV2D maintains a set of keyframes as it processes the video. It estimates optical flow between all pairs of keyframes, then the optical flow is used to solve for a set of pose updates over all pairs jointly. This is done using a differentiable least squares solver which minimizes the reprojection error.

Like DeepV2D, we train on a combination of the NYU and ScanNet datasets using 4 frame video sequences as input. Pose estimation in DeepV2D was trained using an indirect proxy loss on the optical flow induced by the estimated poses. We train using the more direct geodesic error (Eq. \ref{eq:geodesic}), made possible by our library. 

We provide results in Tab. \ref{table:SLAM} and compare to the classical method RGBD-SLAM\cite{kerl2013dense} along with deep learning methods DeepTAM\cite{zhou2018deeptam} and DeepV2D\cite{teed2018deepv2d}. We see that the new loss, difficult to implement with standard backpropagation but made easy by our approach, resulted in improved tracking performance. 

\vspace{-1mm}
\section{Conclusions}
\vspace{-1mm}
We have proposed a new approach for backpropagation through computation graphs containing 3D transformation groups. We have shown that our approach can benefit a wide range of vision and robotics tasks.

\vspace{1mm}
\noindent \textbf{Acknowledgments: } This research is partially supported by a grant (IIS-1942981) from the National Science Foundation and a grant from Samsung.  

{\small
\bibliographystyle{ieee_fullname}
\bibliography{egbib}
}

\appendix











\onecolumn


\begin{center}
    \Large \textbf{Tangent Space Backpropagation for 3D Transformation Groups Appendix}
\end{center}

\section{SO(3), SE(3) and Sim(3) Formulas}
\label{sec:formulas}
Given a Lie group $\mathcal{G}$ with Lie algebra $\mathfrak{g}$, we provide the expressions for $SO(3)$, $SE(3)$, and $Sim(3)$

\vspace{1mm} \noindent \textbf{$\wedge$ Operator:} The $\wedge$ operator takes elements from $\mathbb{R}^k$ to the lie algebra $\mathfrak{g}$. For $\boldsymbol \phi \in \mathbb{R}^3$
\begin{equation}
    \boldsymbol  \phi^\wedge = \begin{pmatrix} 0 & \text{-}\phi_z & \phi_y \\ \phi_z & 0 & \text{-}\phi_x \\ \text{-}\phi_y & \phi_x & 0 \end{pmatrix} \in \mathfrak{so}(3)
\end{equation}
for $\boldsymbol \xi = (\boldsymbol \tau, \boldsymbol \phi) \in \mathbb{R}^6$
\begin{equation}
    \boldsymbol  \xi^\wedge = \begin{pmatrix} 
    \boldsymbol \phi^\wedge & \boldsymbol \tau \\
    0 & 1 \end{pmatrix} = \begin{pmatrix}
    0 & \text{-}\phi_z & \phi_y & \tau_x\\ 
    \phi_z & 0 & \text{-}\phi_x & \tau_y \\ 
    \text{-}\phi_y & \phi_x & 0 & \tau_z \\
    0 & 0 & 0 & 1 \\
    \end{pmatrix} \in \mathfrak{se}(3)
\end{equation}
and for $\boldsymbol \eta = (\boldsymbol \tau, \boldsymbol \phi, \sigma) \in \mathbb{R}^7$
\begin{equation}
    \boldsymbol \eta^\wedge = \begin{pmatrix} 
    \boldsymbol \phi^\wedge + \sigma\mathbf{I}_{3\times3}& \boldsymbol \tau \\
    \sigma & 1 \end{pmatrix} = \begin{pmatrix} 
    \sigma & \text{-}\phi_z & \phi_y & \tau_x\\ 
    \phi_z & \sigma & \text{-}\phi_x & \tau_y \\ 
    \text{-}\phi_y & \phi_x & \sigma & \tau_z \\
    0 & 0 & 0 & 1 \\
    \end{pmatrix} \in \mathfrak{sim}(3)
\end{equation}

\vspace{1mm} \noindent \textbf{$\curlywedge$ Operator:} The $\curlywedge$ operator takes elements from $\mathbb{R}^k$ to the lie algebra $\adj(\mathfrak{g})$, where $\adj(\mathfrak{g})$ is the Lie algebra associated with the group $\Adj(\mathcal{G}) = \{ \Adj(X) \ | \ X \in \mathcal{G} \}$, It can be shown that $\Adj(\mathcal{G})$ also forms a Lie group \cite{barfoot2017state}.

For $\boldsymbol \phi \in \mathbb{R}^3$
\begin{equation}
\boldsymbol  \phi^\curlywedge = \boldsymbol\phi^\wedge \in \adj(\mathfrak{so}(3))
\end{equation}
for $\boldsymbol \xi = (\boldsymbol \tau, \boldsymbol \phi) \in \mathbb{R}^6$
\begin{equation}
\boldsymbol  \xi^\curlywedge = \begin{pmatrix} \boldsymbol \phi^\wedge & \boldsymbol \tau^\wedge \\ \mathbf{0} & \boldsymbol \phi^\wedge \end{pmatrix} \in \adj( \mathfrak{se}(3) ) 
\end{equation}
and for $\boldsymbol \eta = (\boldsymbol \tau, \boldsymbol \phi, \sigma) \in \mathbb{R}^7$
\begin{equation}
\boldsymbol  \eta^\curlywedge = \begin{pmatrix} \boldsymbol \phi^\wedge + \sigma \mathbf{I}_{3\times3} & \boldsymbol \tau^\wedge & -\boldsymbol \tau  \\ \mathbf{0} & \boldsymbol \phi^\wedge & \mathbf{0} \\ \mathbf{0} & \mathbf{0} & 0\end{pmatrix} \in \adj(\mathfrak{sim}(3))
\end{equation}

\vspace{1mm} \noindent \textbf{$\Exp$ Map:} The exponential map takes elements from the Lie algebra to the Lie group. For $SO(3)$, $SE(3)$, and $Sim(3)$ the exponential map has a closed form expressions. For $\boldsymbol \phi \in \mathbb{R}^3$
\begin{equation}
    \Exp(\boldsymbol \phi) = \exp(\boldsymbol \phi^\wedge) = \mathbf{I}_{3\times3} + \frac{\sin(\theta)}{\theta}\boldsymbol\phi^\wedge + \frac{1-\cos(\theta)}{\theta^2} (\boldsymbol\phi^\wedge)^2, \qquad \theta = ||\boldsymbol\phi||
\end{equation}
for $\boldsymbol \xi = (\boldsymbol \tau, \boldsymbol \phi) \in \mathbb{R}^6$
\begin{align}
&\Exp(\boldsymbol \xi) = \begin{pmatrix} \mathbf{R} & \mathbf{V}\boldsymbol\tau \\ \mathbf{0} & 1 \end{pmatrix}, \qquad \mathbf{R} = \Exp(\boldsymbol\phi) \\
&\mathbf{V} = \mathbf{I}_{3\times3} + \frac{1-\cos(\theta)}{\theta^2}\boldsymbol\phi^\wedge + \frac{\theta-\sin(\theta)}{\theta^3} (\boldsymbol\phi^\wedge)^2, \qquad \theta = ||\boldsymbol\phi||
\end{align}
and for $\boldsymbol \eta = (\boldsymbol \tau, \boldsymbol \phi, \sigma) \in \mathbb{R}^7$
\begin{align}
\Exp(\boldsymbol \xi) =& \begin{pmatrix} e^\sigma \mathbf{R} & \mathbf{W}\boldsymbol\tau \\ \mathbf{0} & 1 \end{pmatrix}, \qquad \mathbf{R} = \Exp(\boldsymbol\phi) \\
\mathbf{W} =& \left( \frac{e^\sigma-1}{\sigma} \right) \mathbf{I}_{3\times3} + \frac{1}{\theta}\left( \frac{e^\sigma\sin(\theta)\sigma+(1-e^\sigma\cos(\theta))\theta}{\sigma^2 + \theta^2} \right) \boldsymbol\phi^\wedge + \\  &\frac{1}{\theta^2}\left(\frac{e^\sigma-1}{\sigma} + \frac{(e^\sigma\cos(\theta)-1)\sigma+e^\sigma\sin(
\theta)\theta}{\sigma^2 + \theta^2} \right) (\boldsymbol\phi^\wedge)^2, \qquad \theta = ||\boldsymbol\phi||
\label{eqn:W}
\end{align}
When $\theta$ or $\sigma$ is small, we use second order Taylor approximations of the exponential maps to avoid numerical issues.

\vspace{1mm} \noindent \textbf{$\Log$ Map:} The logarithm map takes elements from the Lie group to the Lie algebra. For $SO(3)$, $SE(3)$, and $Sim(3)$ the logarithm map can be computed in closed form. For a rotation $\mathbf{R} \in SO(3)$
\begin{equation}
    \Log(\mathbf{R}) = \frac{\psi (\mathbf{R} - \mathbf{R}^T)^\vee}{2\sin(\psi)}, \ \psi = \cos^{-1}\left(\frac{tr(\mathbf{R})-1}{2}\right)
\end{equation}
for $\mathbf{G} =\begin{pmatrix} \mathbf{R} & \mathbf{t} \\ \mathbf{0} & 1 \end{pmatrix} \in SE(3)$ we have
\begin{align}
    &\boldsymbol\xi= \begin{pmatrix} \boldsymbol\tau \\ \boldsymbol\phi \end{pmatrix} = \begin{pmatrix} \mathbf{V}^{-1}\mathbf{t} \\ \Log(\mathbf{R}) \end{pmatrix} = \Log(\mathbf{G})\\
    &\mathbf{V}^{-1} = \mathbf{I}_{3\times 3} - \frac{1}{2} \phi^\wedge + \left( \frac{1}{\phi^2} - \frac{1+\cos\theta}{2\theta \sin\theta} \right) (\phi^\wedge)^2, \qquad \theta = ||\boldsymbol \phi||
\end{align}
and for $\mathbf{T} =\begin{pmatrix} s \mathbf{R} & \mathbf{t} \\ \mathbf{0} & 1 \end{pmatrix} \in Sim(3)$
\begin{equation}
\boldsymbol\eta= \begin{pmatrix} \boldsymbol\tau \\ \boldsymbol\phi \\ \sigma \end{pmatrix} = \begin{pmatrix} \mathbf{W}^{-1}\mathbf{t} \\ \Log(\mathbf{R}) \\ \ln(s) \end{pmatrix} = \Log(\mathbf{T})
\end{equation}
where $\mathbf{W}^{-1}$ can be computed by taking the inverse Eqn. \ref{eqn:W}. 

\vspace{1mm} \noindent \textbf{$\Adj$ Operator:} The ajoint operator is a linear map which allows us to move an element $\nu \in \mathfrak{g}$ in the right tangent space of $X \in \mathcal{G}$ to the left tangent space
\begin{equation}
    \Exp(\Adj_X \nu) \circ X = X \circ \Exp(\nu)
\end{equation}
For $\mathbf{R} \in SO(3)$
\begin{equation}
    \Adj_\mathbf{R} = \mathbf{R}
\end{equation}
for $\mathbf{G} =\begin{pmatrix} \mathbf{R} & \mathbf{t} \\ \mathbf{0} & 1 \end{pmatrix} \in SE(3)$
\begin{equation}
    \Adj_{\mathbf{G}} = \begin{pmatrix} \mathbf{R} & \boldsymbol\tau^\wedge\mathbf{R} \\ \mathbf{0} & \mathbf{R} \end{pmatrix}
\end{equation}
and for $\mathbf{T} =\begin{pmatrix} s \mathbf{R} & \mathbf{t} \\ \mathbf{0} & 1 \end{pmatrix} \in Sim(3)$ we have
\begin{equation}
    \Adj_{\mathbf{T}} = \begin{pmatrix} s\mathbf{R} & \boldsymbol\tau^\wedge\mathbf{R} & -\mathbf{t} \\ \mathbf{0} & \mathbf{R} & \mathbf{0} \\ \mathbf{0} & \mathbf{0} & 1 \end{pmatrix}
\end{equation}

\vspace{1mm} \noindent \textbf{$\Inv$ Operator:} Since $SO(3)$, $SE(3)$, and $Sim(3)$ all form a group, each element has a unique inverse.
For $\mathbf{R} \in SO(3)$
\begin{equation}
    \mathbf{R}^{-1} = \mathbf{R}^T
\end{equation}
for $\mathbf{G} =\begin{pmatrix} \mathbf{R} & \mathbf{t} \\ \mathbf{0} & 1 \end{pmatrix} \in SE(3)$
\begin{equation}
    \mathbf{G}^{-1} = \begin{pmatrix} \mathbf{R}^T & -\mathbf{R}^T\mathbf{t} \\ \mathbf{0}  & 1 \end{pmatrix}
\end{equation}
and for $\mathbf{T} =\begin{pmatrix} s \mathbf{R} & \mathbf{t} \\ \mathbf{0} & 1 \end{pmatrix} \in Sim(3)$ we have
\begin{equation}
    \mathbf{T}^{-1} = \begin{pmatrix} s^{-1}\mathbf{R}^T & -s^{-1}\mathbf{R}^T\mathbf{t} \\ \mathbf{0}  & 1 \end{pmatrix}
\end{equation}

\section{Differentials and Jacobians}

\begin{figure}
    \centering
    \includegraphics[width=.5\columnwidth]{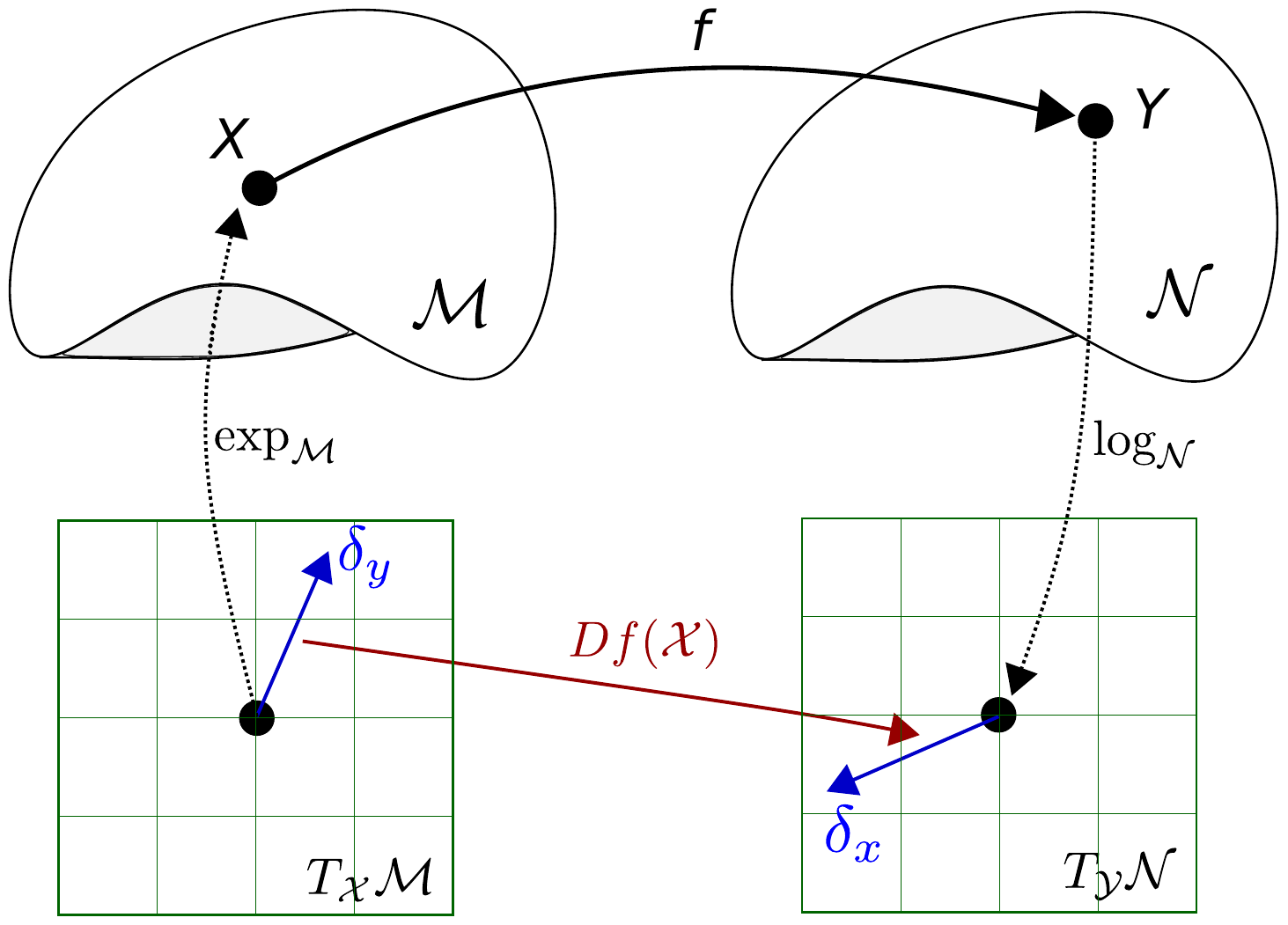}
    \caption{Illistration of the differential between two Lie groups.}
    \label{fig:my_label}
\end{figure}

In the main paper, we derived the gradients for group multiplication. Here we provide derivations of the gradients for the remaining operators

\vspace{1mm} \noindent \textbf{Group Inverse:} Using the definition of the differential 
\begin{align}
D f(X)[\mathbf{v}] &=\lim_{t\rightarrow 0} \frac{\Log((e^{t\mathbf{v}}X)^{-1}(X^{-1})^{-1})}{t} \\
    & = \lim_{t\rightarrow 0} \frac{\Log(X^{-1}e^{-t\mathbf{v}}X)}{t}
\end{align}
using the adjoint
\begin{align}
    & = \lim_{t\rightarrow 0} \frac{\Log(\Exp(-\Adj_{X^{\text{-}1}}(t\mathbf{v})) X^{-1}X)}{t} \\
    & = \lim_{t\rightarrow 0} \frac{-\Adj_{X^{\text{-}1}}t\mathbf{v}}{t} = -\Adj_{X^{\text{-}1}} \mathbf{v}
\end{align}
This can be used to recover the Jacobian $-\Adj_{X^{\text{-}1}}$.

\vspace{1mm} \noindent \textbf{Action on a Point:} We can use elements from the 3D transformation groups to transform a 3D point or set of points. Given a homogeneous point $p=(X, Y, Z, 1)^T$ we can transform $p$ using a transformation $X$
\begin{equation}
    \mathbf{p}' = X\mathbf{p}
\end{equation}
To make the notation consistent for all groups, a rotation can be represented as the $4\times4$ matrix $X  = \begin{pmatrix} \mathbf{R} & 0 \\ 0 & 1 \end{pmatrix}$. $X$ is a linear operator on $p$, so the Jacobian is simply the matrix representation of $X$
\begin{equation}
    \frac{\partial \mathbf{p}'}{\partial \mathbf{p}} = X
\end{equation}
We can also get the differential with respect to the transformation
\begin{align}
    D f(X)[\mathbf{v}] &= \left. \frac{d}{dt} (e^{t\mathbf{v}} X \mathbf{p}) \right |_{t=0} = \left. \frac{d}{dt} (e^{t\mathbf{v}} \mathbf{p}') \right |_{t=0} = \mathbf{v}^\wedge \mathbf{p}'
\end{align}

\vspace{1mm} \noindent \textbf{Adjoint:} We consider the adjoint as the function $Adj: \mathcal{G} \times \mathfrak{g} \mapsto \mathfrak{g}$, $Adj_{X}(\boldsymbol \omega) = \boldsymbol\upsilon$. We need the Jacobians with respect to both $X$ and $\omega$. Since the adjoint is a linear map in terms of $\upsilon$ then
\begin{equation}
    \frac{\partial \boldsymbol\upsilon}{\partial \boldsymbol\omega} = \Adj_X, \qquad \frac{\partial L}{\partial \omega} = \frac{\partial L}{\partial \upsilon} \Adj_X
\end{equation}
where $\Adj_X \in \mathbb{R}^{6\times6}$ is the matrix representation of the adjoint. The gradient with respect to $X$ can be found
\begin{align}
    D \Adj_X(\boldsymbol\omega)[\mathbf v] =& \left. \frac{\partial}{\partial t} \left(e^{t\mathbf{v}}X \boldsymbol \omega^\wedge (e^{t\mathbf{v}} X)^{-1}\right)^\vee \right|_{t=0}\\
    =& \left. \frac{\partial}{\partial t} \left(e^{t\mathbf{v}}X \boldsymbol\omega^\wedge X^{-1} e^{-t\mathbf{v}} \right)^\vee  \right|_{t=0}\\
    =& \left. \frac{\partial}{\partial t} \left(e^{t\mathbf{v}} \boldsymbol\upsilon^\wedge e^{-t\mathbf{v}}  \right)^\vee\right|_{t=0}\\
    =& \left. \left( \frac{\partial}{\partial t} e^{t\mathbf{v}} \boldsymbol\upsilon^\wedge e^{-t\mathbf{v}}  \right)^\vee\right|_{t=0}\\
     =& \left(\mathbf{v}^\wedge\boldsymbol\upsilon^\wedge - \boldsymbol\upsilon^\wedge\mathbf{v}^\wedge\right)^\vee = 
     \boldsymbol\upsilon^\curlywedge \mathbf{v}
\end{align}
Where $\curlywedge$ is defined in Sec \ref{sec:formulas}. We note that the differential is linear in $\mathbf{v}$ allowing us to write the Jacobian and gradients as
\begin{equation}
    \frac{\partial \boldsymbol\upsilon}{\partial X} = \boldsymbol\upsilon^\curlywedge, \qquad \frac{\partial L}{\partial X} = \frac{\partial L}{\partial \upsilon} \boldsymbol\upsilon^\curlywedge
\end{equation}

\vspace{1mm} \noindent \textbf{Exponential and Logarithm Maps:} The Jacobian of the exponential map $\mathbf{J}_l = \frac{\partial }{\partial \mathbf{x}} \Exp(\mathbf{x})$
is referred to as the left-Jacobian and can be written using the series \cite{barfoot2017state} (page 235)
\begin{equation}
    \mathbf{J}_l(\phi) = \sum_{n=0}^{\infty} \frac{1}{(n+1)!} (\phi^\curlywedge)^n
    \label{eqn:left_jacobian}
\end{equation}
For $SO(3)$ and $SE(3)$ closed form expressions exist for Eqn. \ref{eqn:left_jacobian}, otherwise we use the first 3 terms.

The Jacobian of the logarithm map $\mathbf{J}_l^{-1} = \frac{\partial }{\partial X} \Log(X)$ and can be computed using the series
\begin{equation}
    \mathbf{J}_l^{-1}(\phi) = \sum_{n=0}^{\infty} \frac{B_n}{n!} (\phi^\curlywedge)^n
\end{equation}
where $B_n$ are the Bernoulli numbers \cite{barfoot2017state}(page 234). Again, we used analytic expressions of $\mathbf{J}_l^{-1}$ for $SO(3)$ and $SE(3)$, and the first 3 terms for $Sim(3)$.

\section{Sim(3) Network Architecture}

\begin{figure}
    \centering
    \includegraphics[width=.8\columnwidth]{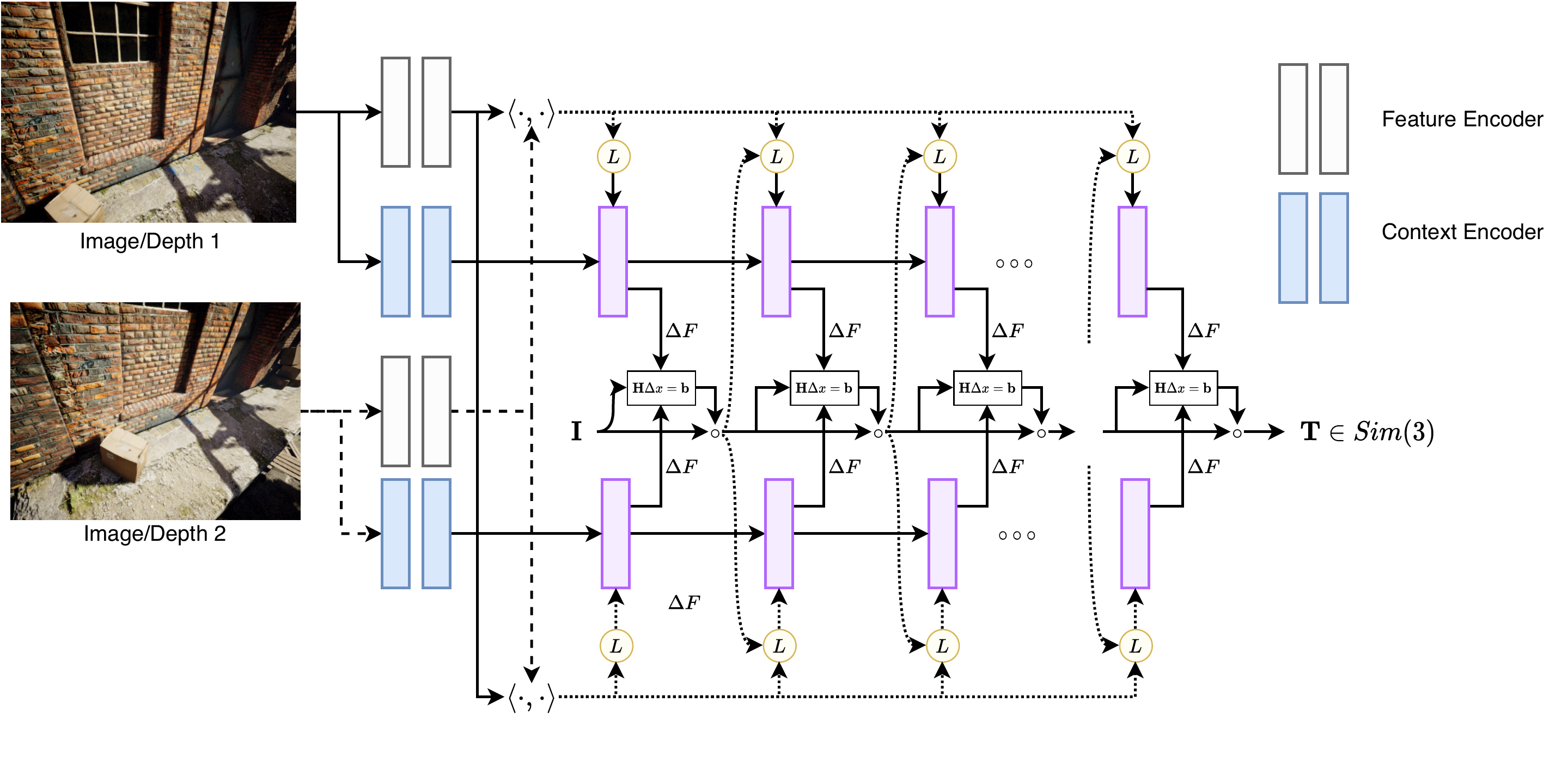}
    \caption{Network architecture used for our Sim(3) registration experiments. The network architecture is based on RAFT\cite{raft}. The top branch estimates motion from $I_1\rightarrow I_2$ and the bottom branch estimates motion in the opposite direction $I_2\rightarrow I_1$. Features are first extracted from each of the two input images and used to construct two 4D correlation volumes, which are pooled at multiple resolutions according to RAFT. During each iteration, the current estimate of the transformation $\mathbf{T}$ is used to index from each of the correlation volumes. The correlation features are processed by the GRU which outputs a residual flow estimate (optical flow not explained by the current transformation $\mathbf{T}$. Both bidirectional residual flow estimates are used as input to a optimization layer, which unrolls 3 Gauss-Newton iterations to produced a transformation update $\Delta \mathbf{T}$, which is applied to the current transformation estimate.}
    \label{fig:sim3_arch}
\end{figure}

A overview of the Sim(3) network architecture is shown in Fig. \ref{fig:sim3_arch}. The context and feature encoders are identical to RAFT. We replace the $5\times1$, $1\times5$ GRU used in RAFT with a single $3\times3$ convolutional GRU, using a hidden state size of 128 channels. We apply 12 iterations during both training and testing.



\end{document}